# Ladder Bottom-up Convolutional Bidirectional Variational Autoencoder for Image Translation of Dotted Arabic Expiration Dates

**Ahmed Zidane[1], BSc**
Ahmed.Zidane.ext@orange.com
Data Scientist
dept. Electronics & Communication
Tanta University

**Ghada Soliman[2]\*, PhD**
Ghada.Soliman@orange.com
Data Scientist Lead
dept. Environmental Engineering
Ain Shams University

**Corresponding Author[2]**
**Orange Innovation Egypt**

**Abstract**

This paper proposes an approach of Ladder Bottom-up Convolutional Bidirectional Variational Autoencoder (LCBVAE) architecture for the encoder and decoder, which is trained on the image translation of the dotted Arabic expiration dates by reconstructing the Arabic dotted expiration dates into filled-in expiration dates. We employed a customized and adapted version of Convolutional Recurrent Neural Network CRNN model to meet our specific requirements and enhance its performance in our context, and then trained the custom CRNN model with the filled-in images from the year of 2019 to 2027 to extract the expiration dates and assess the model performance of LCBVAE on the expiration date recognition. The pipeline of (LCBVAE+CRNN) can be then integrated into an automated sorting systems for extracting the expiry dates and sorting the products accordingly during the manufacture stage. Additionally, it can overcome the manual entry of expiration dates that can be time-consuming and inefficient at the merchants. Due to the lack of the availability of the dotted Arabic expiration date images, we created an Arabic dot-matrix True Type Font (TTF) for the generation of the synthetic images. We trained the model with unrealistic synthetic dates of 60,000 images and performed the testing on a realistic synthetic date of 3000 images from the year of 2019 to 2027, represented as *yyyy/mm/dd*. In our study, we demonstrated the significance of latent bottleneck layer with improving the generalization when the size is increased up to 1024 in downstream transfer learning tasks as for image translation. The proposed approach achieved an accuracy of 97% on the image translation with using the LCBVAE architecture that can be generalized for any downstream learning tasks as for image translation and reconstruction.

# 1 Introduction

Expiration date recognition is a critical problem in the medical and food industries, where the health and safety of consumers depend on the accuracy and efficiency of the detection system. The consequences of consuming expired products can be severe, ranging from mild discomfort to life-threatening illnesses. Therefore, it is crucial to develop a reliable and efficient system that can detect and remove expired products from the shelves before they reach consumers.

Digit recognition is a fundamental problem in computer vision, with many real-world applications such as optical character recognition and automated document processing. While there has been significant progress in digit recognition for Latin characters, recognizing Arabic digits poses a unique challenge due to the complex nature of Arabic script. Moreover, training data for Arabic digit recognition is limited especially for Arabic dot True Type Font (TTF) matrix, and existing datasets often lack diversity, particularly with respect to variations in writing style and quality.

---

Numerous studies have been conducted to explore and advance the field of digit recognition and more specifically on the expiration date recognition, with researchers delving into various aspects of the subject matter to gain deeper insights and address existing challenges. Studies conducted recently have indicated that neural networks exhibit promising performance results when it comes to recognizing expiration dates.

Gong et al. [1] propose a pipeline to detect and recognize the expiration date for an automatic expiration date recognition system. Firstly, the expiration date is detected by extracting the region of interest (ROI) using deep neural network. Following, Image preprocessing techniques with Maximally Stable Extremal Regions (MSER), Component Connected Analysis, and Canny edge detection are applied to make a binarization of the extracted date region with characters being differentiated from the background, identification of the blobs representing different characters, and then extraction of the boundaries of the digits respectively. Tesseract OCR is then employed to segment the digits. Finally, the extracted shapes of the digits are then classified by the nearest neighbor method. The pipeline runs on filled-in images with Latin digits and Color image formats (color/ grayscale).

Muresan, Szabo, and Nedevschi [2] develop a pipeline to detect and recognize expiration dates on water bottles products. The image acquisition was employed using a camera that is positioned in a controlled environment that does not permit light reflection to capture the snapshot of the bottle. At first, the pipeline segments the bottle image using Mask R- CNN [3] with cropping the bottle using the coordinates of the bounding boxes. Next, Image preprocessing techniques are conducted to extract the ROI of the expiration date by resizing and converting to grayscale image, ap- plying morphological gradient operation and binary thresholding using Otsu's Algorithm [4], and then detecting closed contours of the expiration date. The characters are then segmented with adopting equations for horizontal and vertical projections by finding the gaps between characters and resolving the issue of the connected digits. The authors employed the post-processing for reconstructing the dot-matrix characters using dilation of filter 3x3 for 2 iterations with OpenCV. to fill in the missing parts of the digits. The authors proposed a modification on LeNet-5 [5], convolutional neural network architecture to adapt to the single channel of the gray scaled image before the recognition of the segmented digits being taken and the predicted labels are used to identify the corresponding digits. The pipeline runs on filled-in images with Latin digits and grayscale image format on the digits recognition.

Rebedea and Florea [6] propose an end-to-end solution for the detection and recognition of the expiration dates. The authors used TextBoxes++ [7] architecture based on deep neural network to extract regions of interest that might contain expiration dates. The convolutional recurrent neural network (CRNN) is then fine-tuned with the cropped ROI to detect and decode the digits from the expiration date. Finally, a series of regular expressions and logical criteria were carried out following by using a library that can parse time and date in the popular formats. The authors used a dataset that consists of both real images; SynthText [8] and ICDAR [9] with the expiration dates printed on products and synthetically images that are generated by download- able dot matrix type characters with PIL package and then Unity3D graphics to blend in the generated images into the uneven surface of the object. The pipeline runs on filled-in images with Latin digits and colored image format.

Ashino and Takeuchi [10] adopt a pipeline of a combination of two deep neural networks for the detection and recognition of expiration dates on drink packages. The object detection is used to detect the region of interest of the expiration date and recognize the characters (digits and delimiters). The character-recognition DNN is then employed to recognize the characters from these images after being clipped. The pipeline runs on Latin dot matrix characters and colored image format.

Khan [11] proposes a convolutional neural network (CNN) model for the recognition of the expiration date digits with converting the pixel data type from integer to floating-point. The author created a dataset of 1000 pictures where it includes 10 types of digits from 0 to 9. It is comprised of 100 images per each digit. The digits are then re- sized and cropped to 32x32 pixels. The CNN model runs on filled-in images with Latin digits and colored image format.

Gong et al. [12] proposes a pipeline for the detection and recognition of the expiration dates on food package images. The authors adopted a fully convolution neural network for extracting the expiration date followed by the CRNN for the recognition of the digits. The CNN model runs on filled-in images with Latin digits and colored image format.

Seker and Ahn [13] propose a framework of three steps for the detection and recognition of the expiration dates on product packages. The authors used FCOS [14], which was originally developed for object detection by training the model with dates images to detect and extract the expiration date region from an input image. The authors are then adapted FCOS [14] by removing the FPN to reduce network complexity in the DMY detection network, to detect the day, month, and year components from the extracted expiration date region.

Finally, the authors adapted the decoupled attention network (DAN) [15], originally developed for scene and handwritten text recognition, to recognize the characters of the detected day, month, and year regions. As the DAN model was mainly trained with the scene and handwritten text images, the authors performed a fine-tuning of the model with a dataset of synthetic date images created with several expiration date font types and 13 date formats of day, month, and year. The framework runs on Latin dot matrix characters and colored image format.

Our paper proposes a pipeline for the recognition of Arabic dot-matrix characters (digits and delimiters) images on synthetic images of expiration dates in the format of *yyyy/mm/dd*. We adopted Ladder Convolutional Bidirectional Variational Autoencoder (LCBVAE) architecture with Bottom-up for each of the encoder and decoder for Image Translation. The image translation takes the Arabic dot-matrix image as an input, resulting in an output of the corresponding filled-in image. Using the PIL package, we generated synthetic images of dot-matrix characters from an Arabic True Type Font that includes digits 0-9 with varying widths but of uniform and same height, as well as a delimiter symbol ("/"). A font with different widths for the digits can add a unique and visually appealing touch to the design and help with the generalization of the model by increasing the variability of the input dot-matrix images. Alongside the synthetic images, we also created the targeted filled-in images in Cairo font style. We implemented a refined and modified version of Convolutional Recurrent Neural Network (CRNN) model which has been modified to align with our specific use case and maximize its effectiveness within our domain. The custom CRNN is then trained with the targeted filled-in images from the year of 2019 to 2027 to detect and decode the characters from the reconstructed images during inference. Our pipeline produces an accuracy of 97% on the image translation with using the LCBVAE model and can be trained with different resolutions of the dot-matrix without modifying the resolution of the corresponding filled-in image whilst still producing the same performance results.

Our study has also proven that LCBVAE architecture with bottom-up for the encoder-decoder obtained better results for the image translation of the Arabic dot matrix image in terms of accuracy and training time compared to the conventional autoencoder where it is comprised of up-down for the encoder and bottom-up for the decoder. Moreover, it also shows that a larger latent space can lead to a better generalization performance in a variational autoencoder (VAE) as it captures more complex relationship between the input images and the encoded representations in the form of gaussian multivariate distribution. The output of the VAE is then generated by sampling from the probability distribution over the latent space, rather than by decoding a fixed encoding. This is done using a decoder network that takes a sample from the latent space as input and generates a reconstructed output. Figure 1 shows our pipeline that is comprised of LCBVAE and our custom CRNN.

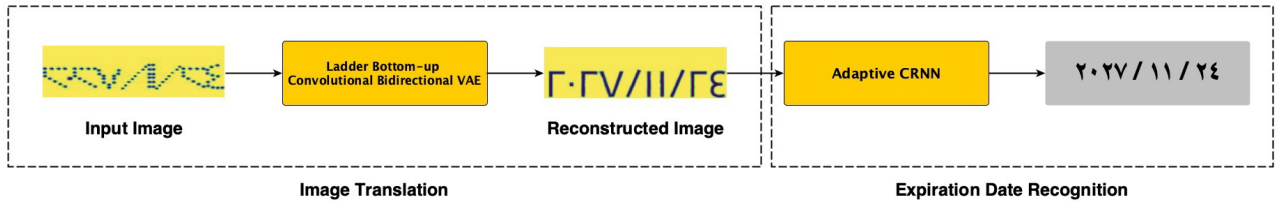

Figure 1: Expiration Date Pipeline

## 2 Dataset Generation

### 2.1 Challenges

**Lack of Real Data** - In the Arabic-speaking world, the lack of a standardized format for the expiration date on food and medical products poses a significant challenge for consumers and retailers alike. The Arabic expiration date can be written in various formats. There are also no public datasets for Arabic dot-matrix digits that support variations in the fonts and styles and allows to generalize effectively across various writing styles and contexts.

**Traditional Filling Methods** - Traditional erosion and dilation techniques have been widely used to fill in dotted digits in various languages for digit recognition tasks. However, when it comes to Arabic dotted digits, these techniques have proven to be ineffective for our custom synthetic dataset. This is mainly because the spacing between Arabic digits in our dataset is almost zero, which makes it challenging for traditional erosion and dilation techniques to accurately reconstruct the dots as shown in Figure 2 .

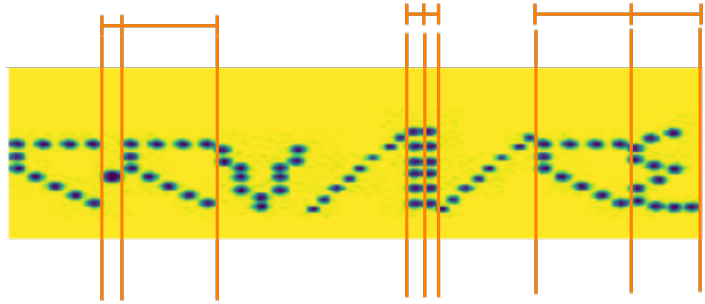

Figure 2: Challenges with Irregular Spacing among Characters

## 2.2 Generate Synthetic Data using Arabic Dot-matrix TTF

Due to the lack of public dataset on Arabic dot-matrix format, the synthetic dataset is generated using Arabic dot-matrix TTF where the characters are drawn as vector graphics and then saved as TrueType Font (TTF) using FontForge as shown in Figure 3.

## 2.3 Training and Testing Dataset

The dataset consists of 60,000 samples of unrealistic expiry dates with the corresponding filled-in expiry dates that incorporates more samples for training the model. A larger dataset with varied placements of digits helps the LCBVAE) model learn a more robust representation of the data. It ensures that the model can generalize well to unseen data, capturing the underlying distribution more effectively. There are 3000 samples of realistic dates dataset covering the years 2019 to 2027, used for testing the model. Samples of Realistic and Unrealistic dates is shown in Figure 4.

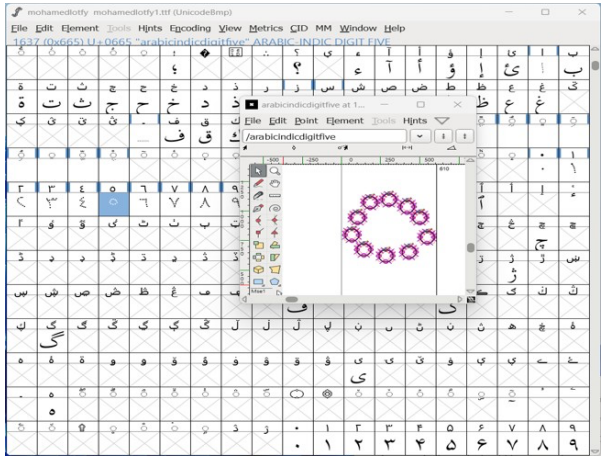

Figure 3: Font-Forge

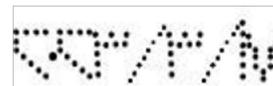 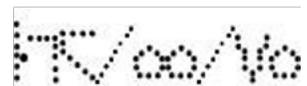

(a) Realistic Date  (b) Unrealistic Date

Figure 4: Realistic and Unrealistic Date Samples

# 3 Methodology

## 3.1 Motivation

Variational Autoencoder (VAE) is a type of deep generative model that is used to learn low-dimensional representations of high-dimensional data as shown in Figure 5 [16]. VAE enhance the conventional autoencoders by adopting a probabilistic framework to learn a latent representation of the input data. Instead of directly encoding the input data into a low-dimensional representation, VAEs first learn a probability distribution over the latent variables that describe the underlying structure of the input data. The model then samples from this learned distribution to generate a latent representation of the input data.

Generalization in Variational Autoencoders (VAEs) refers to the ability of the model to perform well on unseen data that is not present in the training set. In this study, we aim to decode the expiry date from Arabic dot-matrix images. To achieve this, we trained the LCBVAE model as shown in Figure 6 on unrealistic Arabic dates. This ensures that the model detects variations of the number positions to reconstruct dates, regardless of the positions of the numbers. This allows the model to make it robust against variations in number positions. However, factors such as lighting and rotation are not considered in this study. The following model architecture is supposed to reconstruct images from the dot-matrix to filled-in format by training on unrealistic Arabic date samples in the format of *yyyy/mm/dd* as for example: 9999/99/9.

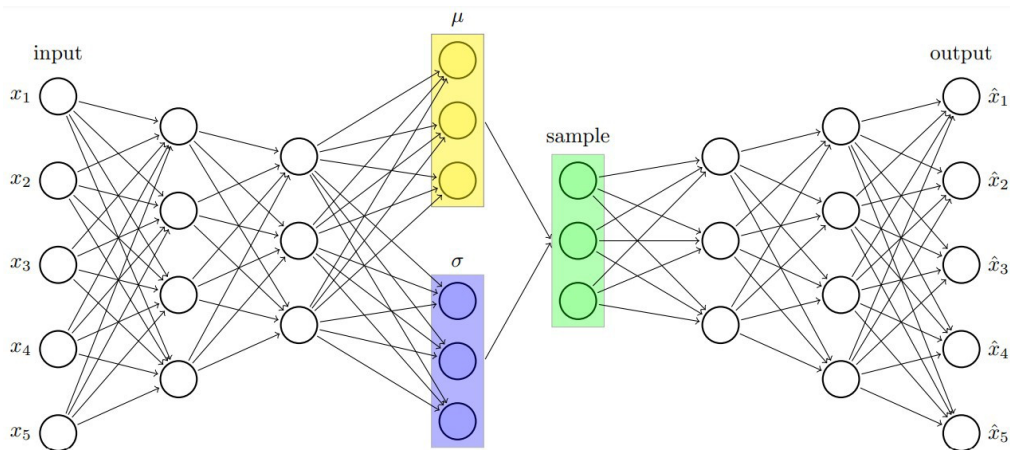

Figure 5: Variational Autoencoder Reprinted from Source

## 3.2 LCBVAE Model Architecture

**LCBVAE architecture**, as shown in Figure 6 consists of three main parts Encoder, Latent, and Decoder as illustrated in the following subsections.

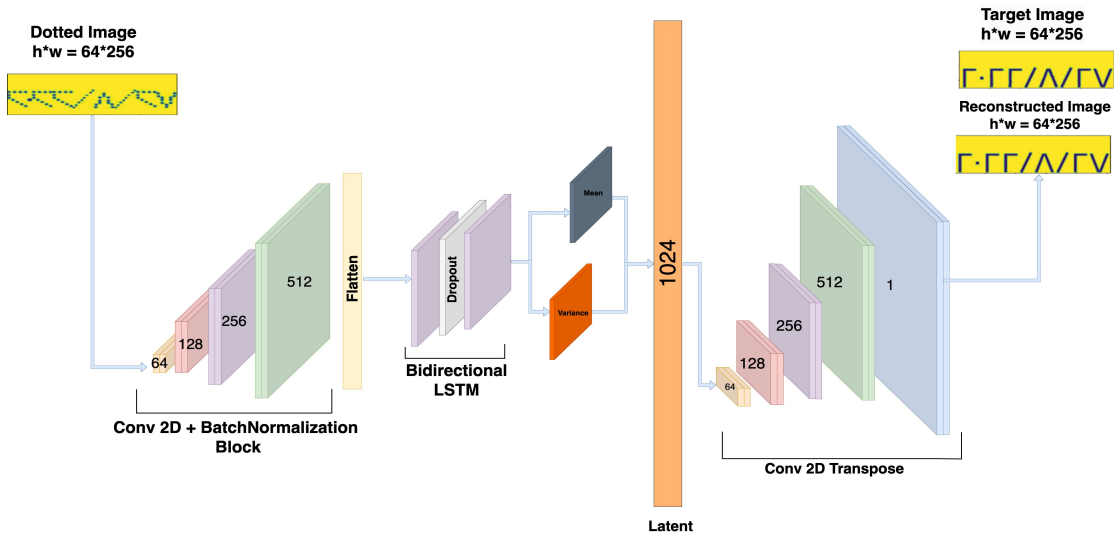

Figure 6: LCBVAE Architecture

### 3.2.1 Encoder (Recognition Model) Architecture of Variational Autoencoder

In this section, we present the architecture of the encoder component of a Variational Autoencoder (VAE) model, as illustrated in Figure 7. The encoder takes an input image $X$ of size (64, 256, 1) and applies a series of convolutional layers to extract features from the input image. The encoder is comprised of convolutional layers with filter sizes of 64, 128, and 256, followed by batch normalization. In our work, the pooling and unpooling are not used in the model architecture of VAE, as they may discard useful image details that are essential for the reconstruction task [17]. In contrast to high-level applications such as segmentation or recognition, pooling typically eliminates abundant image details and may deteriorate restoration performance [17]. The encoder also incorporates bidirectional layers, culminating in a sampling layer that generates the latent space representation. The resulting compressed latent space representation of the input image is denoted as $H_{enc}$ and is used to generate two vectors: the mean vector, denoted by $z_{mean}$, and the variance vector, denoted by $z_{logvar}$. These vectors define a latent space that is used as input to the decoder component for generating the reconstructed images. The details of the encoder architecture are shown in Table 1.

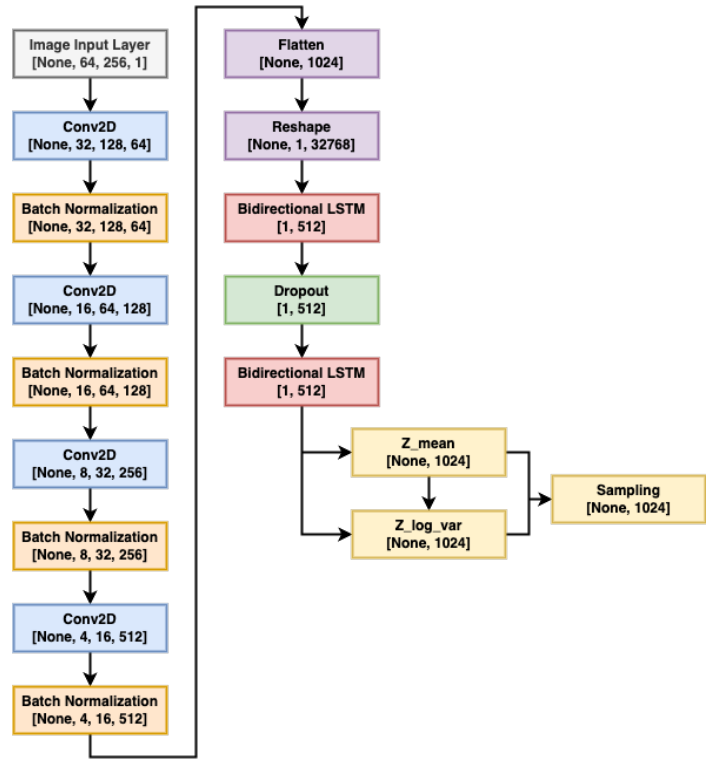

Figure 7: Encoder Architecture

In a Variational Autoencoder (VAE), the encoder is designed to represent a probabilistic distribution over the latent variables rather than a single deterministic point. This is achieved by parameterizing the encoder to output the parameters of a probability Gaussian distribution. These parameters are represented by the mean and variance of the latent variables that determine its properties from the data through a neural network. The process of sampling from a distribution that is parameterized by the encoder is not differentiable. Hence, the reparameterization trick is applied to make the network differentiable by adding an independent noise term $\epsilon$ that is sampled from typically a normal distribution with mean zero and standard deviation one. This Gaussian sample can then be scaled by the predicted mean and variance that produce samples drawn from a fixed Gaussian distribution enabling the model to cover unseen samples in the input data [16].

Formally, the encoder function $f_{enc}$ applies a series of convolutional and pooling layers followed by batch normalization to obtain the compressed representation $H_{enc}$ from the input image $X$, and ReLU as an activation function [18], which can be expressed mathematically as:

$$H_{enc} = f_{enc}(X) \tag{1}$$

The compressed representation $H_{enc}$ is then used to compute the mean and variance vectors of the latent space, given by:

$$z_{mean} = W_{mean}H_{enc} + b_{mean} \tag{2}$$

$$z_{logvar} = W_{logvar}H_{enc} + b_{logvar} \tag{3}$$

Where $W_{mean}$, $b_{mean}$, $W_{logvar}$, and $b_{logvar}$ are learnable weights and biases of the bidirectional layer in the encoder.

Finally, the encoder produces a sample from the latent space by computing a reparameterization trick using the mean and variance vectors. The sample Z is then used as input to the decoder component for generating novel images.

$$Z = z_{mean} + \epsilon \odot e^{z_{logvar}/2} \tag{4}$$

Where ε is a random variable drawn from a standard normal distribution, and ⊙ denotes elementwise multiplication. This trick regularizes the latent space [16].

Table 1: Summary of the Encoder Model

| Layer Type | Output Shape | Param # |
| --- | --- | --- |
| InputLayer | (None, 64, 256, 1) | 0 |
| Conv2D | (None, 32, 128, 64) | 640 |
| BatchNormalization | (None, 32, 128, 64) | 256 |
| Conv2D | (None, 16, 64, 128) | 73,856 |
| BatchNormalization | (None, 16, 64, 128) | 512 |
| Conv2D | (None, 8, 32, 256) | 295,168 |
| BatchNormalization | (None, 8, 32, 256) | 1,024 |
| Conv2D | (None, 4, 16, 512) | 1,180,160 |
| BatchNormalization | (None, 4, 16, 512) | 2,048 |
| Flatten | (None, 32768) | 0 |
| Reshape | (None, 1, 32768) | 0 |
| Bidirectional | (None, 1, 512) | 67,635,200 |
| Dropout | (None, 1, 512) | 0 |
| Bidirectional | (None, 256) | 656,384 |
| mean | (None, 1024) | 263,168 |
| Variance | (None, 1024) | 263,168 |
| Sampling | (None, 1024) | 0 |

### 3.2.2 Latent layer

The latent layer is a crucial component of the Variational Autoencoder (VAE) architecture [16]. The purpose of the latent layer is to learn a low-dimensional representation of the data that captures its essential features. The size of the latent layer is a hyperparameter that needs to be chosen before training the VAE. The latent layer's size determines the dimensionality of the low-dimensional representation that the VAE learns. In practice, the size of the latent layer is usually chosen to be much smaller than the input data's dimensionality. This constraint ensures that the VAE learns a compact representation that captures the essential features of the input data.

We experimented with different latent sizes, including 32, 64, 128, 256, 512, and 1024. However, we found that a larger latent size of 1024 provided the best performance. This result suggests that a bigger latent size can improve the model's generalization and robustness.

Reparameterization Trick For continuous latent variables and a differentiable encoder and generative model, the ELBO can be straightforwardly differentiated with respect to both φ and θ through a change of variables, also called the reparameterization trick [16].

### 3.2.3 Decoder (Generative Model)

This section presents the decoder architecture that is used in our model. The architecture consists of a series of transposed convolutional layers with ReLU activation function and a final sigmoid activation layer to produce the reconstructed image. The decoder takes a low-dimensional representation of the input data generated by the encoder and then generates a reconstructed version of the original input, denoted by **z**.

The latent representation z is a sample drawn from the approximate posterior distribution over the latent variables p(z|x) where x is the input data. The decoder, as shown in Figure 8, consists of four transposed convolutional layers with 64, 128, 256, and 512 filters respectively, each followed by ReLU activation function. The first three transposed convolutional layers have a stride of 2, which increases the spatial resolution of the feature maps while decreasing the number of filters. The final transposed convolutional layer has a stride of 1 to maintain the spatial resolution of the feature maps. The decoder outputs a reconstructed image, denoted by x̃, which is generated by passing the final feature map through a sigmoid activation function. The reconstructed image x̃ is a continuous-valued matrix with the same dimensions as the original input.

Table 2 lists the hyper-parameters used in the decoder architecture, including the number of filters in each transposed convolutional layer, the size of the filters, and the stride of the first three transposed convolutional layers.

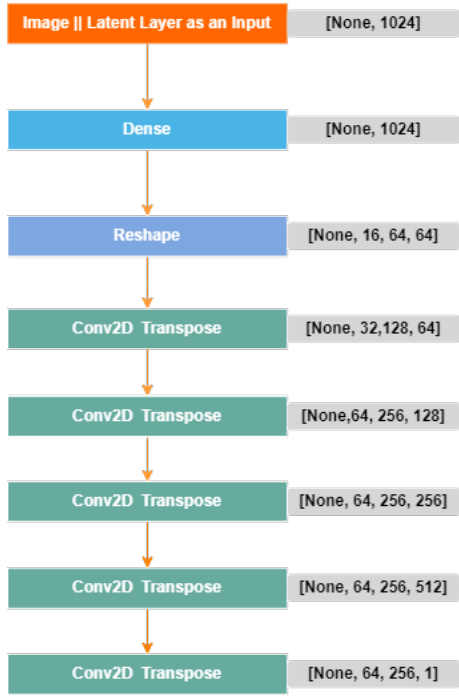

Figure 8: Decoder using Transposed Convolutional Layer (TCL)

Table 2: Decoder Model Architecture with Transposed Convolutional Layer

| Layer Type | Output Shape | Param # |
| --- | --- | --- |
| input 2 | (InputLayer) | 0 |
| dense | (None, 65,536) | 67,174,400 |
| reshape 1 (Reshape) | (None, 16, 64, 64) | 0 |
| conv2d transpose | (None, 32, 128, 64) | 36,928 |
| conv2d transpose 1_ | (None, 64, 256, 128) | 73,856 |
| conv2d transpose 2_ | (None, 64, 256, 256) | 295,168 |
| conv2d transpose 3_ | (None, 64, 256, 512) | 1,180,160 |
| conv2d transpose 4_ | (None, 64, 256, 1) | 4,609 |

### 3.2.4 Transposed Convolutional Layer (TCL)

In our study, we used TCL [19] instead of downsampling of the decoder. The key difference between TCL and downsampling layers lies in their respective operations. While downsampling layers typically use pooling or striding operations to reduce the spatial resolution of the feature maps, TCL use a learnable transpose convolution operation to increase the spatial resolution of the feature maps. The transpose convolution operation works by reversing the forward and backward passes of a regular convolution operation. During the forward pass, the transpose convolution operation performs a convolution between the input feature map and a set of learnable filters, while during the backward pass, it performs an upsampling operation that increases the spatial resolution of the feature map.

## 4 Reconstruction and Regularization Loss

In this section, we discuss the loss function used in variational autoencoders (VAEs). The VAE loss is known as the Evidence Lower Bound (ELBO) or the variational lower bound. It provides a tractable objective function for training VAEs that incorporates both reconstruction loss and the regularization term. The ELBO [16] is defined as follows:

$$L(\theta, \emptyset; x) = \frac{1}{L} \sum_{j=1}^{L} \log p_\theta(x, z_j) - \log q_\emptyset(z_j|x) \qquad (5)$$

Where $\theta$ and $\phi$ represent the parameters of the generative model $p_\theta(x, z)$ and the inference model $q_\phi(z|x)$, respectively. *L* denotes a minibatch of data samples.

The first term in the ELBO is the reconstruction loss, which measures the negative log-likelihood of the data given the latent variables. It encourages the generative model to produce reconstructions that resemble the original data points. The second term in the ELBO is the Kullback-Leibler (KL) [20] divergence between the approximate posterior distribution $q_\phi(z|x)$ and the prior distribution $p(z)$. This term acts as a regularizer, promoting the disentanglement of latent representations and encouraging the approximate posterior to match the prior distribution.

During training, the VAE optimizes the ELBO by computing its gradients with respect to the parameters $\theta$ and $\phi$ and updating them using an SGD optimizer [16]. This iterative process continues until convergence, resulting in learned parameters that capture the underlying data distribution and enable generation of new samples. By maximizing the ELBO, the VAE finds a balance between reconstructing the data and regularizing the latent space, leading to meaningful and expressive latent representations. The variational autoencoder loss plays a crucial role in training VAEs and is a key component in learning powerful generative models.

## 5 Custom CRNN

Shi et al. [21] implemented an end-to-end Trainable Neural Network for Scene Text Recognition whose network architecture as shown in Table 3 is specifically designed for recognizing sequence-like objects in images. It is comprised of three components, including the convolution layer that extracts a sequence of feature vectors from the feature maps, which is then used as an input into a deep bidirectional recurrent neural network that predicts the label distribution of each frame in the feature sequence. The transcription is the final layer that predicts the label sequence per frame that has the highest probability.

We employed a customized version of the CRNN model as shown in Figure 9 to meet our specific requirements and enhance its performance in our context with referring to Table 4. We were inspired by the conventional encoder component of the autoencoder architecture when redesigning the convolution layers in our model for two folds: (1) decreasing the feature maps aids in reducing the computational cost of the network. By progressively reducing the spatial dimensions and the number of feature maps, the model requires fewer parameters and computations, making it more computationally efficient, and (2) reducing the number of feature maps helps to capture and summarize the essential information from the input data, discarding less relevant or redundant details. This compression of information promotes a more efficient representation learning process, where the model focuses on the most salient features for the task.

Table 3: Reference CRNN Configurations

| Type | Configurations |
|---|---|
| Bi-LSTM | 2 * (hidden units:256) |
| Convolution | maps:512, k:2 * 2, s:1, p:0 |
| MaxPooling | Window:1 x 2, s:2 |
| Convolution | 2 * (maps:512, k:3 * 3, s:1, p:1) |
| MaxPooling | Window:1 x 2, s:2 |
| Convolution | 2 * (maps:512, k:3 * 3, s:1, p:1) |
| MaxPooling | Window:2 x 2, s:2 |
| Convolution | maps:128, k:3 * 3, s:1, p:1 |
| MaxPooling | Window:2 x 2, s:2 |
| Convolution | maps:64, k:3 * 3, s:1, p:1 |
| Input | W x 32 gray-scale image |

Table 4: Custom CRNN Configurations

| Type | Configurations |
|---|---|
| Bi-LSTM | 3 * (hidden units:16) |
| Convolution | maps:4, k: 3 * 3, s:1, p:1 |
| MaxPooling | Window: 2 x 2, s:2 |
| Convolution | maps:8, k: 3 * 3, s:1, p:1 |
| MaxPooling | Window: 2 * 2, s:2 |
| Convolution | maps:16, k: 3 * 3, s:1, p:1 |
| Input | 256 * 64 * 3 colored image |

Table 5: Adaptive vs Reference RCNN Configurations, (a): Custom CRNN Configurations, (b): Reference CRNN Configurations

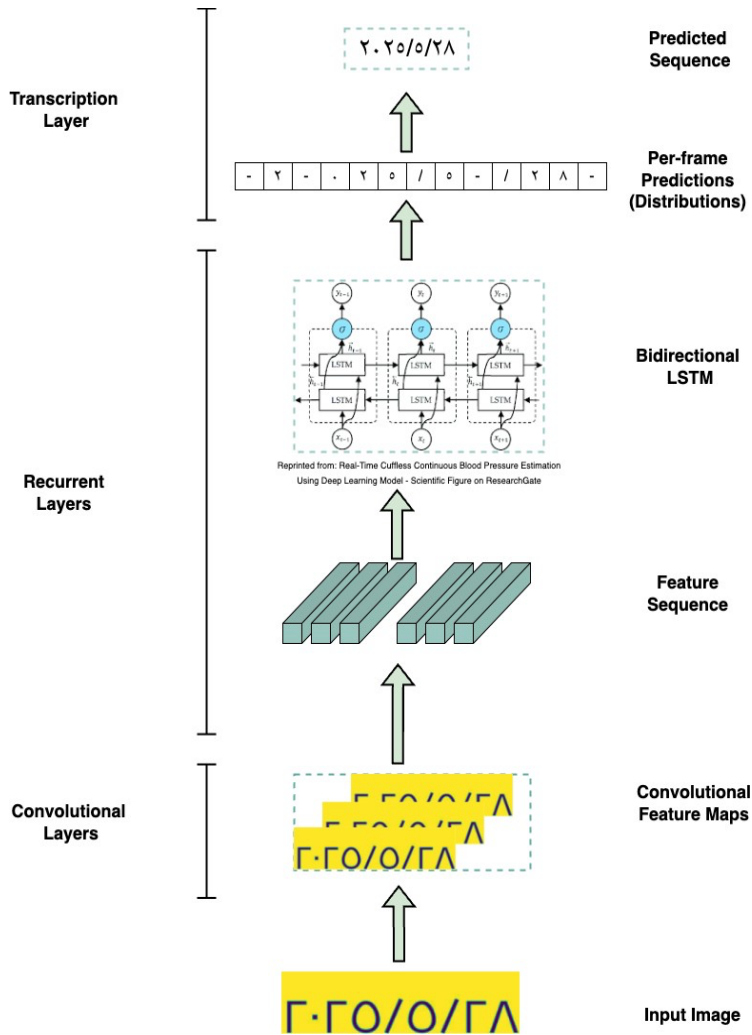

Figure 9: Our Custom CRNN Model

## 5.1 Connectionist Temporal Classification (CTC) Loss

The Connectionist Temporal Classification (CTC) technique, which was initially introduced by Graves et al. [22], has emerged as a prominent approach in Optical Character Recognition (OCR) tasks. This technique addresses the challenge of aligning variable-length input and output sequences in OCR by allowing flexible mapping between image sequences and text sequences. It achieves this through the incorporation of a special" blank" label and a mechanism for label repetition. In CTC, a probability distribution over labels is generated by the network at each time step during training. The CTC algorithm then determines the most likely alignment, considering possible repetitions of labels and insertions of blank labels. This approach enables accurate recognition of text from images, even in the presence of misalignment or variation in sequence lengths. By effectively handling variable-length input and output sequences, the CTC technique has become a valuable tool in OCR, enabling the accurate extraction of text information from images.

## 6 Results

This section presents the results of the pipeline models (LCBVAE+CRNN). The models are implemented in Keras and have been run on P100 Nvidia GPU with 16 GB RAM.

**Our findings** as shown in Table 6 demonstrate that the combination of Bidirectional Long Short-Term Memory (LSTM) Hochreiter and Schmidhuber [23] architecture with 1024 latent units, absence of pooling layers, and utilization of dropout regularization, yielded the most favorable outcomes. The model exhibited stability after 20 epochs, although 50 epochs were required to reach an acceptable loss threshold. During the initial 0 to 18 epochs, the reconstruction loss experienced a significant reduction from an initial value of 4000 to 1350. Subsequently, from epoch 18 to 50, the loss exhibited a more gradual decline from 1350 to 1100. These results underscore the effectiveness of the proposed configuration in improving the model's performance and convergence. Notably, the model successfully translated or reconstructed the dotted font into a solid, discernible font, rendering it easily detectable by any Optical Character Recognition (OCR) system.

Additionally, we explored the impact of dataset size and batch size on the model's performance. With a dataset size of exactly 60,000 images and a batch size of 32, our results were deemed satisfactory as shown in Table 7. Figure 10 demonstrates the progression of the reconstructed images during the training phase of epochs from 1 to 18. The input image represents an Arabic dot-matrix font Expiry date Image, while the reconstructed Image denotes the output produced by our model. Our primary objective was to maximize the similarity between the reconstructed image and the target image. These findings highlight the progressive refinement of our model's ability to accurately reconstruct the target image throughout the training process.

The presented Figure 11 depicts the relationship between the loss and the number of epochs. Each epoch has an approximate duration of 10 minutes, and the training process encompasses approximately 1900 samples. Each sample consists of a pair of images, where the input image represents the Arabic dot-matrix font, and the output image represents the corresponding Arabic reconstructed image.

**Our Custom CRNN** is trained for 50 epochs with the targeted filled in images from the year of 2019 to 2027 to detect and decode the characters from the reconstructed images during inference. Using our Custom CRNN model, we achieved a significant improved loss value of 0.098, compared to a loss of 2.9697 when using the original CRNN configurations [21] without any modifications. Moreover, our customized model exhibits a significantly reduced number of trainable parameters, with only 12K, in contrast to the original CRNN model, which has 8.3M trainable parameters. This reduction in the number of parameters highlights the efficiency and lightweight nature of our model, while still delivering competitive performance. We achieved accuracy of 97% on the test images of 3000 realistic synthetic images from 2019 to 2027.

Table 6: Experiments on Latent Space with Bidirectional LSTM and Dense

| Experiment | Training Dataset | Testing Dataset | Accuracy |
|---|---|---|---|
| With Bidirectional LSTM | Unrealistic 60,000 Sample | Realistic 3000 Sample | 97% |
| With Dense | Unrealistic 60,000 Sample | Realistic 3000 Sample | 92% |

Table 7: Model Summary

| Epochs | Training Time | Inference Time/image | Accuracy | Number of Weights | Model Size |
|---|---|---|---|---|---|
| 18 | 2~3 hours | 6.1 ms | 97% | $140 * 10^6$ | 600 MB |

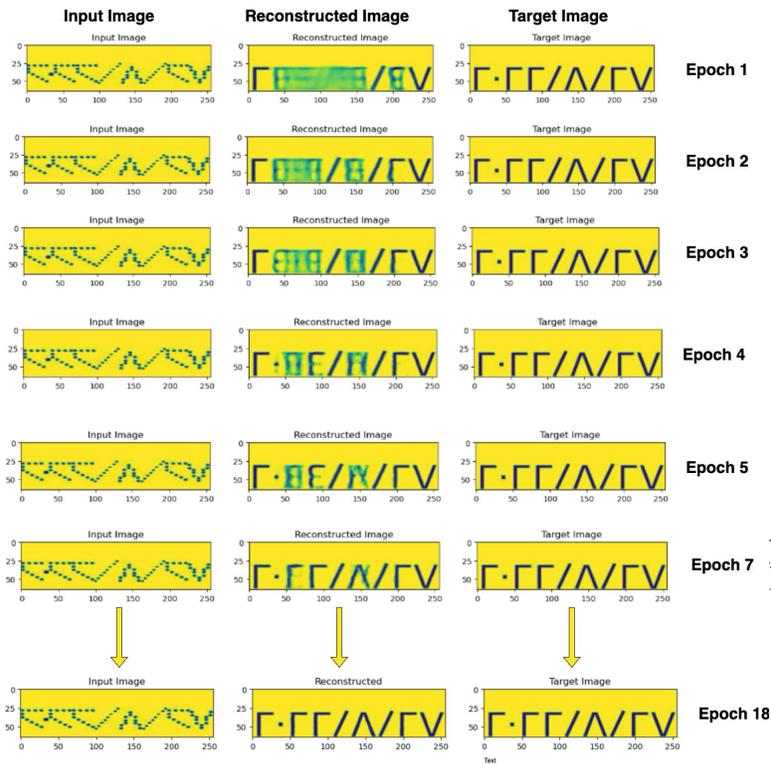

Figure 10: Visualizing Results during Training from epoch 1 to epoch 18

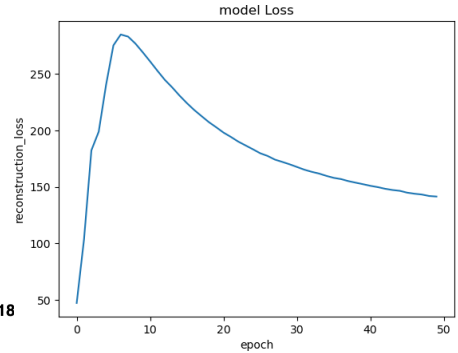

Figure 11: Reconstruction Loss

Table 8 shows a comparison of our work against the recent previous works on the expiry date recognition models. Almost the previous approaches are primarily focused on the expiry date recognition in Latin characters while our approach relies on Arabic dot-matrix digits. Recognizing Arabic digits presents several challenges compared to Latin digits including variations in writing style, size, shape, and slant, as well as image noise [24]. These factors can alter numeral topology, increasing ambiguity in the recognition process of Arabic digits. In addition, our work demonstrates a significant efficiency in processing time with a rate of 18.3 images per second during the inference time in comparison to the inference speed time of the approaches of Rebedea and Florea [6], Khan [11], and Seker and Ahn [13].

Table 8: Summary of Approaches on the Expiry Date Recognition Methods

| Approach | Dataset (Synthetic/ Real) | Image Format | Image Content | Latin/ Arabic | Filled in/ Dot-matrix | Inference Speed (img/s) | Recognition Approach | Test Accuracy (%) |
|---|---|---|---|---|---|---|---|---|
| Rebedea and Florea [6] | Both | Colored | Expiry Date | Latin | Both | 0.25 | RCNN | 72.7 |
| Ashino and Takeuchi [10] | Real | Binarized | Single Digit | Latin | Dot-matrix | - | DNN | 90 |
| Khan [11] | Real | Colored | Single Digit | Latin | Both | 0.89 | CNN | 90 |
| Gong et al [12] | Real | Colored | Expiry Date | Latin | Dot-matrix | - | CRNN | 95 |
| Seker and Ahn [13] | Both | Colored | Expiry Date | Latin | Both | 0.12 | DAN | 97.74 |
| **Ours** | **Synthetic** | **Binarized** | **Expiry Date** | **Arabic** | **Dot-Matrix** | **18.3** | **Image Translation with Variational Autoencoders + CRNN** | **97** |

# 7 Conclusion

Emphasizing the value of decoding Arabic dot-matrix digits becomes crucial due to the lack of existing research or papers on this specific aspect of Optical Character Recognition (OCR). As no prior studies have addressed this subject, our focus on decoding Arabic dot-matrix digits gains significant importance. The absence of relevant literature highlights the novelty and potential impact of this research, underscoring the need to explore and develop robust methods to tackle this challenging problem effectively. By investigating and presenting our findings comprehensively, we aim to contribute substantially to the field of OCR and pave the way for advancements in recognizing Arabic dot-matrix digits.

In our work, we developed a generalized model for reconstructing the Arabic dot-matrix dates images into filled-in images, trained on Arabic unrealistic dates. Our model architecture includes variational Autoencoder, with specific optimization techniques such as dropout and batch-normalization. Our findings indicate that the most favorable outcomes are achieved when utilizing an architecture comprising bottom-up convolutional layers and bidirectional LSTM, while excluding pooling layers. Moreover, we noticed that LSTM performed over the Dense layer in terms of accuracy and reconstruction. Dense layers are faster in the warm-up epochs, but eventually it is failed to reconstruct the middle of the image.

Our VAE was trained with a 1024 latent layer, which allowed for stable reconstructed images after 18 epochs. Our Custom CRNN model, which utilized the CTC loss function, achieved an accuracy of 97% percent in predicting/ decoding the Arabic expiration dates given the translated image output of VAE. We considered the predicated date as misclassified for the presence of any single wrong character.

As a potential avenue for future research, the extension of our proposed approach to image reconstruction in different domains holds significant promise. By exploring the adaptability of the Bidirectional Long Short-Term Memory (LSTM) architecture, along with variations in latent units, loss functions, pooling layers, and dropout regularization, we can assess the effectiveness of our model in diverse image reconstruction tasks. Furthermore, investigating its performance on alternative datasets from various domains would provide valuable insights into its generalizability and applicability. Therefore, considering the application of our approach to image reconstruction in other domains constitutes an important area for future investigation.

In summary, our results demonstrate the effectiveness of our generalized model in reconstructing realistic dates, and the importance of specific optimization techniques and architecture choices in achieving superior performance. Our findings can potentially be applied to a range of other problems requiring image reconstruction and translation.

# References

1. L. Gong, M. Yu, W. Duan, X. Ye, K. Gudmundsson, and M. Swainson, "A Novel Camera Based Approach for Automatic Expiry Date Detection and Recognition on Food Packages." *IFIP Advances in Information and Communication Technology*, pp. 133-142, 2018, doi: 10.1007/978-3-319-92007-8_12.
2. M. P. Muresan, P. A. Szabo and S. Nedevschi, "Dot Matrix OCR for Bottle Validity Inspection," *2019 IEEE 15th International Conference on Intelligent Computer Communication and Processing (ICCP)*, Cluj-Napoca, Romania, 2019, pp. 395-401, doi: 10.1109/ICCP48234.2019.8959762.
3. K. He, G. Gkioxari, P. Dollár and R. Girshick, "Mask R-CNN," *2017 IEEE International Conference on Computer Vision (ICCV)*, Venice, Italy, 2017, pp. 2980-2988, doi: 10.1109/ICCV.2017.322.
4. D. Liu and J. Yu, "Otsu Method and K-means," *2009 Ninth International Conference on Hybrid Intelligent Systems*, Shenyang, China, 2009, pp. 344-349, doi: 10.1109/HIS.2009.74.
5. Y. Lecun, L. Bottou, Y. Bengio and P. Haffner, "Gradient-based learning applied to document recognition," in *Proceedings of the IEEE*, vol. 86, no. 11, pp. 2278-2324, Nov. 1998, doi: 10.1109/5.726791.
6. T. Rebedea and V. Florea, "Expiry date recognition using deep neural networks." *International Joural of User-System Interaction*, vol. 13, no. 1, pp. 1-17, 2020, doi: 10.37789/ijusi.2020.13.1.1.
7. M. Liao, B. Shi, and X. Bai, "TextBoxes++: A Single-Shot Oriented Scene Text Detector." *IEEE Transactions on Image Processing*, vol. 27, no. 8, pp. 3676-3690, 2018, doi: 10.1109/tip.2018.2825107.
8. A. Gupta, A. Vedaldi and A. Zisserman, "Synthetic Data for Text Localisation in Natural Images," *2016 IEEE Conference on Computer Vision and Pattern Recognition (CVPR)*, Las Vegas, NV, USA, 2016, pp. 2315-2324, doi: 10.1109/CVPR.2016.254.
9. A. Shahab, F. Shafait and A. Dengel, "ICDAR 2011 Robust Reading Competition Challenge 2: Reading Text in Scene Images," *2011 International Conference on Document Analysis and Recognition*, Beijing, China, 2011, pp. 1491-1496, doi: 10.1109/ICDAR.2011.296.
10. M. Ashino and Y. Takeuchi, "Expiry-Date recognition system using combination of deep neural networks for visually impaired," in *Lecture notes in computer science*, 2020, pp. 510–516. doi: 10.1007/978-3-030-58796-3_58.
11. T. Khan, "Expiry Date Digit Recognition using Convolutional Neural Network," *European Journal of Electrical Engineering and Computer Science*, vol. 5, no. 1, pp. 85–88, Feb. 2021, doi: 10.24018/ejece.2021.5.1.259.
12. L. Gong, M. Thota, M. Yu, W. Duan, M. Swainson, X. Ye, and S. Kollias. "A novel unified deep neural networks methodology for use by date recognition in retail food package image," *Signal, Image and Video Processing*, vol. 15, no. 3, pp. 449–457, Sep. 2020, doi: 10.1007/s11760-020-01764-7.
13. A. C. Seker and S. C. Ahn, "A generalized framework for recognition of expiration dates on product packages using fully convolutional networks," *Expert Systems With Applications*, vol. 203, p. 117310, Oct. 2022, doi: 10.1016/j.eswa.2022.117310.
14. Z. Tian, C. Shen, H. Chen and T. He, "FCOS: Fully Convolutional One-Stage Object Detection," *2019 IEEE/CVF International Conference on Computer Vision (ICCV)*, Seoul, Korea (South), 2019, pp. 9626-9635, doi: 10.1109/ICCV.2019.00972.
15. T. Wang, Y. Zhu, L. Jin, C. Luo, X. Chen, Y. Wu, Q. Wang, and M. Cai. 2020. Decoupled Attention Network for Text Recognition. *Proceedings of the AAAI Conference on Artificial Intelligence*. vol. 34, no. 07, pp. 12216-12224, 2020.
16. D. P. Kingma and M. Welling, "An introduction to variational autoencoders," *Foundations and Trends in Machine Learning*, vol. 12, no. 4, pp. 307–392, Jan. 2019, doi: 10.1561/2200000056.
17. X.-J. Mao, C. Shen, and Y.-B. Yang, "Image Restoration Using Convolutional Auto-encoders with Symmetric Skip Connections," arXiv preprint arXiv:1606.08921, 2016.
18. L. Bottou, "Stochastic gradient descent tricks," in *Lecture notes in computer science*, 2012, pp. 421–436. doi: 10.1007/978-3-642-35289-8_25.
19. E. Shelhamer, J. Long and T. Darrell, "Fully Convolutional Networks for Semantic Segmentation," in *IEEE Transactions on Pattern Analysis and Machine Intelligence*, vol. 39, no. 4, pp. 640-651, 1 April 2017, doi: 10.1109/TPAMI.2016.2572683.
20. S. Kullback and R. A. Leibler. On Information and Sufficiency. *The Annals of Mathematical Statistics*. 1951. Vol. 22(1):79-86. DOI: 10.1214/aoms/1177729694.
21. B. Shi, X. Bai and C. Yao, "An End-to-End Trainable Neural Network for Image-Based Sequence Recognition and Its Application to Scene Text Recognition" in *IEEE Transactions on Pattern Analysis & Machine Intelligence*, vol. 39, no. 11, pp. 2298-2304, 2017. doi: 10.1109/TPAMI.2016.2646371

22. A. Graves, S. Fernandez, F. Gomez, and J. Schmidhuber. "Connectionist temporal classification: labelling unsegmented sequence data with recurrent neural networks" in *Proceedings of the 23rd international conference on Machine learning*, pp. 369–376, 2006.
23. S. Hochreiter and J. Schmidhuber. Long short-term memory. *Neural Computation*, 9(8):1735–1780, 1997.
24. A. Alani, "Arabic handwritten digit recognition based on restricted Boltzmann machine and convolutional neural networks," *Information*, vol. 8, no. 4, p. 142, Nov. 2017, doi: 10.3390/info8040142.